\documentclass[10pt,twocolumn,letterpaper]{article}

\usepackage{cvpr}              
\definecolor{cvprblue}{rgb}{0.21,0.49,0.74}
\usepackage[pagebackref,breaklinks,colorlinks,allcolors=cvprblue]{hyperref}
\usepackage{hyperref}
\usepackage{url}
\usepackage{amssymb}
\usepackage{pifont}
\usepackage{bbding}
\usepackage{xcolor}
\usepackage{booktabs}
\usepackage{amsmath}
\usepackage{float}
\usepackage{placeins}
\usepackage{colortbl}
\usepackage{courier}
\usepackage{graphicx}
\usepackage{natbib}
\usepackage{caption}
\usepackage{wrapfig}
\usepackage{multirow}
\usepackage[ruled,linesnumbered,noend,vlined]{algorithm2e}

\newcommand{\cmark}{\ding{51}}

\newcommand{\model}{Reason3DVG}

\def\paperID{} 
\def\confName{CVPR}
\def\confYear{2026}

\title{Reasoning Matters for 3D Visual Grounding}

\author{
Hsiang-Wei Huang\quad
Kuang-Ming Chen\quad
Wenhao Chai\quad
Cheng-Yen Yang\quad\\
Jen-Hao Cheng\quad
Jenq-Neng Hwang\quad\\
University of Washington\\
{\tt\small \{hwhuang,kmchen,wchai,cycyang,andyhci,hwang\}@uw.edu}
}

\begin{document}
\maketitle
\begin{abstract}
The recent development of Large Language Models~(LLMs) with strong reasoning ability has driven research in various domains such as mathematics, coding, and scientific discovery. Meanwhile, 3D visual grounding, as a fundamental task in 3D understanding, still remains challenging due to the limited reasoning ability of recent 3D visual grounding models. Most of the current methods incorporate a text encoder and visual feature encoder to generate cross-modal fuse features and predict the referring object. These models often require supervised training on extensive 3D annotation data. On the other hand, recent research also focus on scaling synthetic data to train stronger 3D visual grounding LLM, however, the performance gain remains limited and non-proportional to the data collection cost. In this work, we propose a 3D visual grounding data pipeline, which is capable of automatically synthesizing 3D visual grounding data along with corresponding reasoning process. Additionally, we leverage the generated data for LLM fine-tuning and introduce Reason3DVG-8B, a strong 3D visual grounding LLM that outperforms previous LLM-based method 3D-GRAND using only 1.6\% of their training data, demonstrating the effectiveness of our data and the importance of reasoning in 3D visual grounding.
\end{abstract}
    
\begin{figure}[!t]
    \centering
    \includegraphics[width=0.98\linewidth]{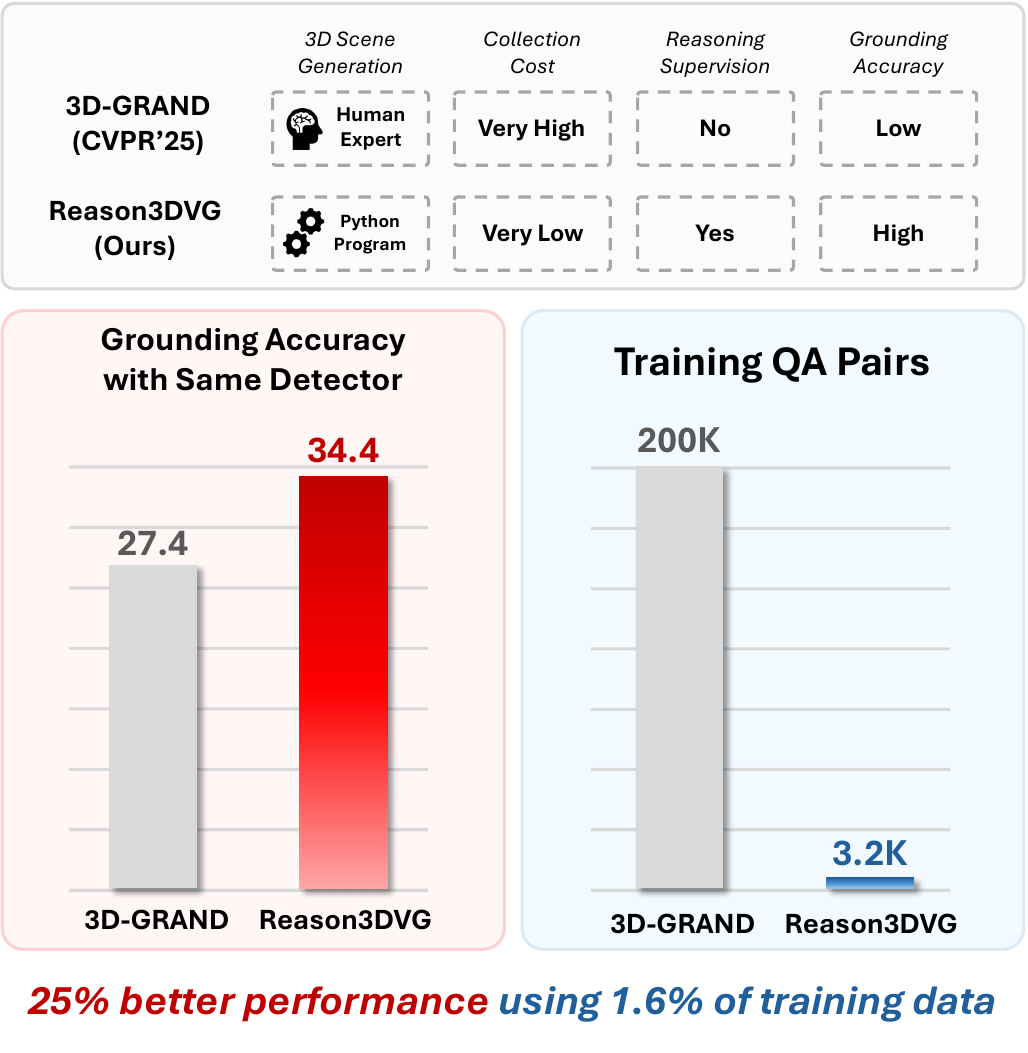}
    \caption{Our method outperforms 3D-GRAND on the ScanRefer benchmark when using only \textbf{1.6\%} amount of their training data scale. Compared with 3D-GRAND, our proposed data pipeline features lower data collection cost, incorporate extra reasoning supervision for LLM, and achieve \textbf{25\%} better grounding accuracy.}
    \label{fig:data}
\end{figure}
\section{Introduction}
\label{sec:intro}
3D Visual Grounding is a fundamental task in 3D understanding, aiming to identify a target object within a 3D scene based on a given textual query. Recent supervised-training models~\cite{butd-detr,concretenet,mcln,g3lq,eda,zhao20213dvg} integrate a text encoder and a visual encoder to generate cross-modal features for target object prediction. These supervised methods rely on large-scale, real-world annotated 3D visual grounding datasets for supervised training. To mitigate this, some recent research also explores approaches that leverage proprietary Large Language Models~(LLMs) and Vision Language Models~(VLMs) for the 3D visual grounding task via an agentic workflow~\cite{llmgrounder,li2024seeground} or through in-context examples and code generation~\cite{visualprogramming}. Despite these methods somewhat demonstrating success in the 3D visual grounding task, they heavily rely on proprietary LLMs and VLMs to achieve the best visual grounding performance, which introduce extensive inference cost during test time.

\begin{figure*}[t]
    \centering
    \includegraphics[width=0.98\linewidth]{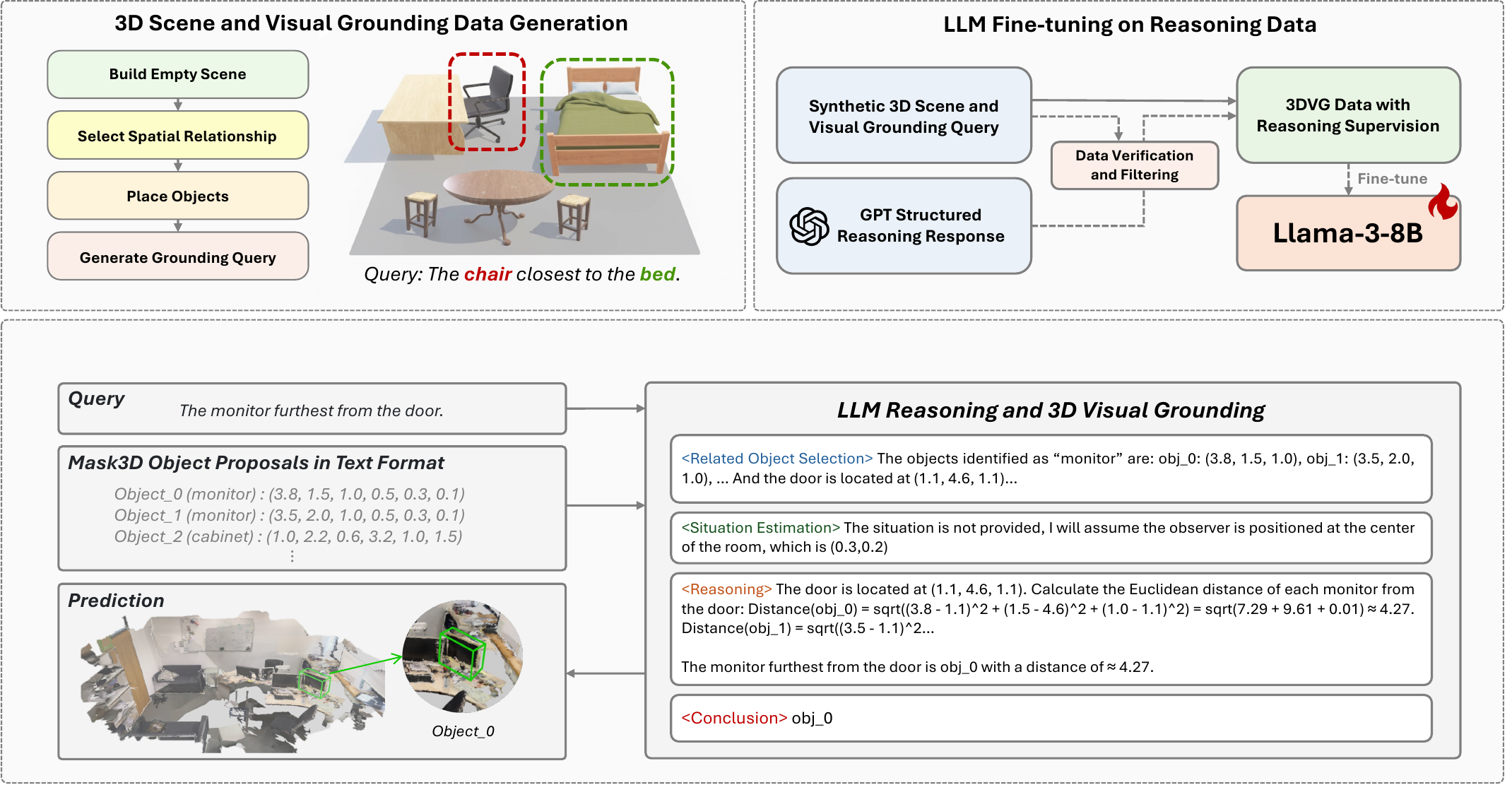}
    \caption{We propose a fully automatic data pipeline that can generate visual grounding queries and reasoning responses. The collected data are used to conduct LLM fine-tuning, which results in \model-8B, a powerful LLM with strong 3D visual grounding ability.}
    \label{fig:teaser}
\end{figure*}

To advance open-source models for improved performance on 3D visual grounding, recent work also investigates the direction of collecting 3D visual grounding data for open-source LLM fine-tuning. 3D-GRAND~\cite{3d-grand} collects a million-scale 3D visual grounding dataset and fine-tunes open-source LLM on this large-scale data. Despite showing performance improvements on public benchmarks, several new challenges emerge. First, 3D-GRAND relies on human-expert-designed 3D scenes, which require extensive manual labor. In addition, the dense object-level annotations introduce further cost. Furthermore, their experimental results also indicate that the performance improvement achieved through fine-tuning on such large-scale data is moderate and not proportional to the significant data collection effort. These challenges motivate us to rethink 1) a more cost-efficient, fully automatic data pipeline that can generate 3D visual grounding data in a human-free and cost-efficient fashion, and 2) the actual key beyond data scale towards improving the performance of open-source LLM on the challenging 3D visual grounding task.

In this work, we explore the direction of improving the LLM's 3D visual grounding performance using fully automatically generated synthetic data, featuring detailed, structured reasoning supervision for LLM fine-tuning. Our proposed automated data collection pipeline introduces minimal cost, featuring detailed, structured reasoning supervision. LLM fine-tuned on our data outperforms previous LLM-based method 3D-GRAND that trained on \textbf{60$\times$} more data, as shown in Fig.~\ref{fig:data}. 

Our work first introduces a fully automatic, 3D visual grounding data pipeline, which aims to address the dependency on extensive human-annotated 3D visual grounding data. We use our collected data and conduct LLM fine-tuning, resulting in \model, an LLM with strong reasoning ability and advanced 3D visual grounding accuracy. We perform extensive evaluation and show that \model~achieves superior grounding accuracy than SOTA zero-shot methods and LLM-based method 3D-GRAND on multiple 3D visual grounding benchmarks including ScanRefer and NR3D. We summarize the main contribution of our work as follows:

\begin{itemize}
    \item We propose a fully automatic 3D visual grounding data pipeline for LLM fine-tuning. Our data pipeline does not require any human annotation, which largely reduces the data collection cost compared to previous works.
    \item We conduct LLM fine-tuning on our collected data and introduce \model, an LLM for 3D visual grounding task that achieves strong accuracy on multiple 3D visual grounding benchmarks including ScanRefer and NR3D.
    \item Our model outperforms previous 3D visual grounding LLM 3D-GRAND when using only 1.6\% of training data, demonstrating the importance of reasoning supervision in the 3D visual grounding task, and serves as a cornerstone for future 3D understanding LLM development with stronger reasoning and understanding ability.
    
\end{itemize}
\section{Related Work}

\begin{figure*}[t]
    \centering
    \includegraphics[width=1\linewidth]{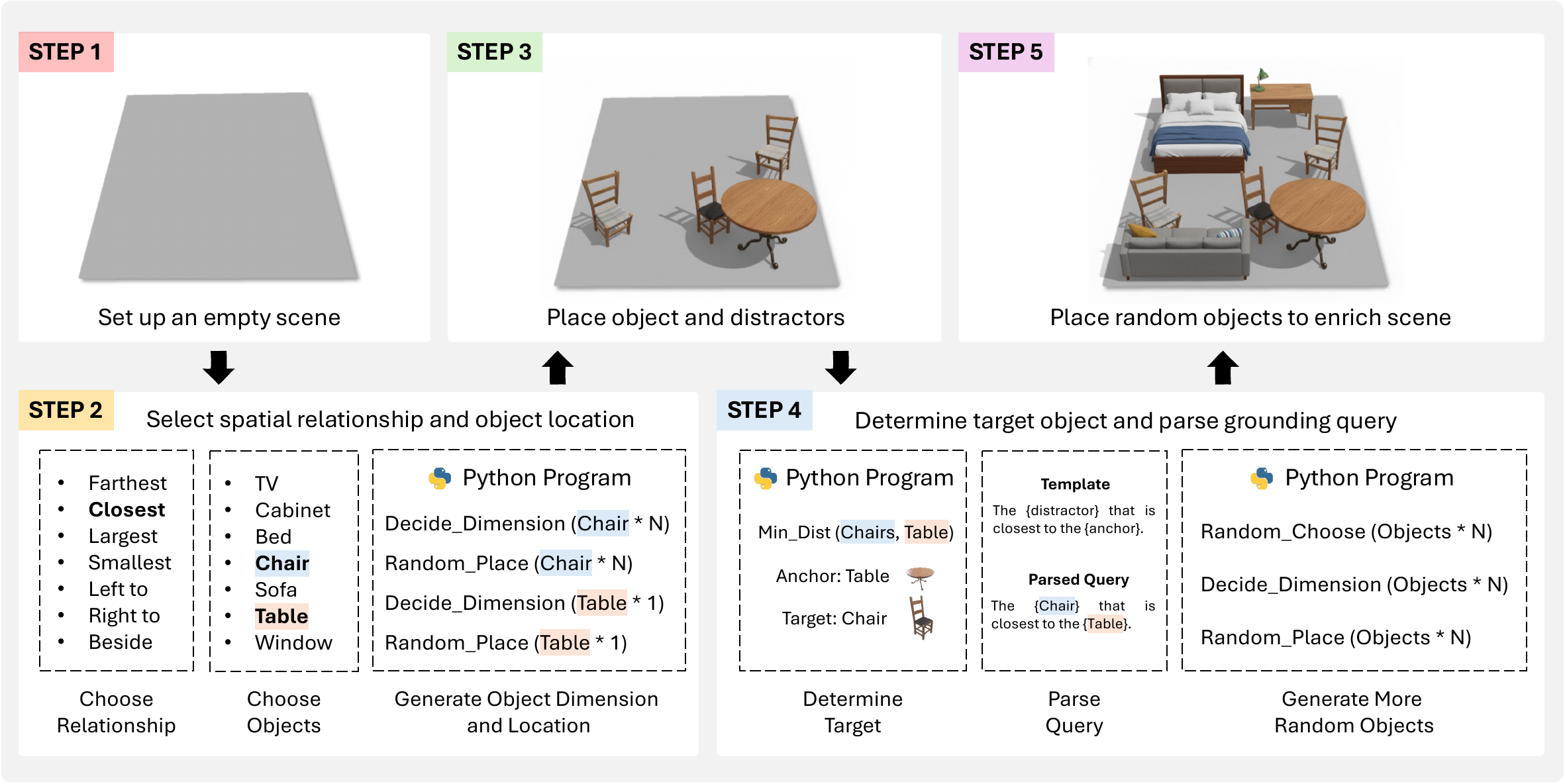}
    \caption{An illustration of our 3D scene layout generation pipeline, which includes 5 steps:~\textbf{1)} Set up an empty 3D scene with certain dimension. \textbf{2)} Choose a spatial relationship and decide the anchor and distractor objects as well as their dimension and location. \textbf{3)} Place them in the 3D scene. \textbf{4)} Determine the target object from the distractors and query. \textbf{5)} Generate more distractor objects to enrich 3D scene.}
    \label{fig:pipeline}
\end{figure*}

3D visual grounding is a fundamental 3D understanding task that aims to predict the target object in the 3D scene based on a given language query. Most of the recent 3D visual grounding methods~\cite{eda,butd-detr,3dsps,MVT} adopt separate encoders for each input modality such as text and point cloud to generate cross-modal features and conduct prediction. Some end-to-end methods like SAT~\cite{SAT}, LanguageRefer~\cite{roh2022languagerefer}, and UniT3D~\cite{chen2023unit3d} utilized a unified multi-modal transformer to conduct 3D visual grounding. Most of these methods utilize supervised training on human collected 3D data, which incur extensive annotation efforts. Recent works started to focus on the LLM-based approaches~\cite{llmgrounder,visualprogramming}, which leverage proprietary LLM to solve the 3D visual grounding task, but introduce extensive inference costs. 3D-GRAND~\cite{3d-grand} attempts to scale-up the training data for small-scale, open-source LLM training, yet it still requires extensive 3D scene layout annotation, and only achieve sub-optimal grounding performance. In this work, we explore using an open-source LLM to solve the 3D visual grounding task through our fully automatic data collection pipeline, which does not require any human expert annotations, and outperforms existing LLM-based methods on multiple 3D visual grounding benchmarks.
\section{Method}
\label{sec:method}
We illustrate our proposed framework in Fig.~\ref{fig:teaser}. We design an automatic 3D visual grounding data pipeline that features detailed, structured reasoning responses for LLM fine-tuning. We fine-tuned open-source LLM Llama-3.1-8B on our collected data. During test time, we utilize an object detector Mask3D to generate object proposals, which are transformed to text format and sent to the LLM along with the query. The LLM then predicts the referring target object by performing multi-stage structured reasoning.

\subsection{3D Scene Generation}
\label{subsec:scene}
The collection of 3D scene data requires a huge effort, including extensive human expert annotations and indoor 3D scan collection. To address this, we propose to generate synthetic 3D scene data for LLM-based model training and enable the model to conduct 3D visual grounding in real-world 3D scenes. There are many existing methods for 3D scene generation~\cite{feng2023layoutgpt,yang2024holodeck}, but they mostly focus on generating scene layout with realistic object arrangements or following the specified room style from users' prompt, while our goal is to collect object-centric 3D scene that focuses on objects' spatial arrangements as well as the target object's corresponding grounding query, which is an important component of 3D visual grounding data.

In this work, we design a program-based 3D scene data pipeline that can generate an object-centric 3D scene following several common spatial relationships between objects~(shown in Fig.~\ref{fig:pipeline}-Step.2). Here, we provide a generation example of a 3D scene layout in Fig.~\ref{fig:pipeline}, which contains the 3D scene layout following query \textit{"The Chair that is closest to the Table."} with $n$ distractor chairs in the scene:
\begin{enumerate}
    \item Define a 3D scene size with a certain dimension $H\times W$.
    \item Place $n$ chairs and one table in the 3D scene.
    \item Calculate the distance between each chair and table to find the closest chair to the table; this chair will be the target object of the 3D visual grounding query.
    \item Generate more objects to enrich the 3D scene diversity.
\end{enumerate}
Following this procedure, we collect a wide range of 3D scene layouts that follow our specified common spatial relationship and are not confined by the limited number of 3D scene layouts in the real-world available 3D scans. For each spatial relationship, we design multiple templates from which we can parse the target's object class to generate the final target object query. We prompt \texttt{GPT-4o} to generate some indoor scene objects as well as their common dimension. When placing objects, we randomly select object classes and apply a random variation on their size to enhance diversity. We attached our code for 3D scene generation, as well as all our generated 3D scenes in the supplemental materials for reference.

It is important to note that our data pipeline \textbf{are not designed to generate complex queries and spatial relationships that simulate the real-world complexity}. This is because recent LLMs already possess basic reasoning ability; our data utilizes several common spatial relationships with detailed reasoning responses serving as supervision to encourage the LLM to think step-by-step in a chain-of-thought manner for the 3D visual grounding task. Furthermore, our experiments and ablation study results show that after fine-tuning on our simple spatial relationships, \textbf{the LLM can solve complex, real-world 3D visual grounding queries that were not seen in our collected data}, as suggested in our study in Tab.~\ref{tab:domain}. Lastly, because our goal is to generate 3D scene data that follows our specified spatial relationship between objects, we only refrain the object from being placed overlapping during scene generation, and do not focus on placing them in a realistic manner.

\subsection{Four Stages Reasoning}
\label{subsec:stages}
Recent advanced LLMs with reasoning ability adopt special tokens or tags~\cite{guo2025deepseek,gemini} to separate the reasoning and conclusion stage, which enables a more structured and organized thinking process of the LLM. In the 3D visual grounding domain, it is also important to think step-by-step in a chain-of-thought manner before generating the final prediction, especially when trying to interpret the complicated spatial relationship in the target query. To achieve this, we proposed four structured reasoning stages, as illustrated in our Fig.~\ref{fig:teaser}. These four chain-of-thought reasoning stages encourage the LLM to generate a more structured thinking process. The four stages include:

\begin{itemize}
    \item \textbf{Related Object Selection.} Given all the objects in the 3D scene, the LLM listed the objects that might be relevant to the given query. Each listed object will include information such as its object ID, class name, location, and dimension.
    \item \textbf{Situation Estimation.} If a 3D situation~(viewer's location) is provided, the LLM will estimate its possible coordinate in the 3D scene. Otherwise, the LLM defaults to assume it is situated in the middle of the scene.
    \item \textbf{Reasoning.} This is the main stage for LLM to conduct reasoning based on the previously given information. The LLM will conduct necessary math calculations and logical reasoning to derive the final answer.
    \item \textbf{Conclusion.} In this stage, the model outputs the final prediction in a structured format, which enables easier parsing, evaluation and comprehension by the user.
\end{itemize}

\subsection{Reasoning Data Collection}
\label{subsec:reason}
To collect detailed reasoning data aligned with our four reasoning stages, we leverage proprietary LLM \texttt{GPT-4o} to generate structured reasoning responses. We design a specialized prompt to guide \texttt{GPT-4o} in adhering to our four-stage reasoning process. The full prompt and the 3D scene input format is provided in the appendix. We send our collected queries and corresponding 3D scenes in structured text format to \texttt{GPT-4o} to generate reasoning response. Each sample includes a unique 3D scene layout with over 50 objects, a target object's language query, its corresponding object ID, and the reasoning response from \texttt{GPT-4o}.

After data collection, we further perform a verification process to ensure the quality of our dataset. Since each 3D scene layout is generated through rule-based code, the layout is \textbf{guaranteed} to correctly correspond to the specified spatial relationships. For the \texttt{GPT-4o} generated responses, we further validate the final predicted answer against our generated ground truth and remove any responses with incorrect predictions. Approximately 10\% of responses that deviate from our specified format or contain incorrect answers are automatically filtered through string pattern matching and answer verification. The remaining verified reasoning responses are incorporated into our final training set, which contains 3.2K samples.

To further measure the correctness of the reasoning process, we also conduct manual verification on a subset of the responses. We observe that most major reasoning steps are correct, with only minor arithmetic errors (typically after the first decimal place) that do not affect the correctness of final predictions. Therefore, we do not apply additional filtering based on intermediate reasoning results and instead use the final answer as our main filtering criterion.

\begin{table*}[!t]
    \centering
    \small
    \caption{Performance comparison with existing methods on ScanRefer validation set. We follow previous work and report the grounding accuracy under different IoU threshold, and also report the accuracy of \textit{``Unique"} (scenes with a single target object) and \textit{``Multiple"} (scenes with distractors of the same class) subsets, along with overall performance.}
    \resizebox{\linewidth}{!}{
    \begin{tabular}{l|c|cc|cc|cc}
        \toprule 
        \multirow{2}{*}{\textbf{Method}} & \multirow{2}{*}{\textbf{Source}} & \multicolumn{2}{c|}{\textbf{Unique}} & \multicolumn{2}{c|}{\textbf{Multiple}} & \multicolumn{2}{c}{\textbf{Overall}} 
        \\
        & & \textbf{Acc@$\mathbf{0.25}$} & \textbf{Acc@$\mathbf{0.5}$} & \textbf{Acc@$\mathbf{0.25}$} & \textbf{Acc@$\mathbf{0.5}$} & \textbf{Acc@$\mathbf{0.25}$} & \textbf{Acc@$\mathbf{0.5}$} 
        \\
        \midrule
        \textit{\textbf{Fine-tuned on Training Set}} \\
        ScanRefer~\cite{scanrefer} & ECCV'20 &  $67.6$ & $46.2$ & $32.1$ & $21.3$ & $39.0$ & $26.1$
        \\
        InstanceRefer~\cite{yuan2021instancerefer} & ICCV'21 & $77.5$ & $66.8$ & $31.3$ & $24.8$ & $40.2$ & $32.9$ 
        \\
        3DVG-Transformer~\cite{zhao20213dvg} & ICCV'21 & $77.2$ & $58.5$ & $38.4$ & $28.7$ & $45.9$ & $34.5$ 
        \\
        BUTD-DETR~\cite{butd-detr} & ECCV'22 & $84.2$ & $66.3$ & $46.6$ & $35.1$ & $52.2$ & $39.8$
        \\
        EDA~\cite{eda} & CVPR'23 & $85.8$ & $68.6$ &  $49.1$  & $37.6$  & $54.6$ &  $42.3$
        \\
        3D-VisTA~\cite{3d-vista} & ICCV'23 & $81.6$ & $75.1$ & $43.7$ & $39.1$  & $50.6$ &   $45.8$ 
        \\
        G3-LQ~\cite{g3lq} & CVPR'24 & $88.6$ & $73.3$ & $50.2$ & $39.7$  & $56.0$ &   $44.7$ 
        \\
        MCLN~\cite{mcln} & ECCV'24 & $86.9$ & $72.7$ & $52.0$ & $40.8$  & $57.2$ &   $45.7$
        \\
        ConcreteNet~\cite{concretenet} & ECCV'24 & $86.4$ & $82.1$ & $42.4$ & $38.4$  & $50.6$ &   $46.5$ 
        \\
        \midrule
        \textit{\textbf{Not Fine-tuned on Training Set}} \\
        LERF~\cite{kerr2023lerf} & ICCV'23 & - & -  & - & - & $4.8$ & $0.9$ 
        \\
        OpenScene~\cite{peng2023openscene} & CVPR'23 & $20.1$ & $13.1$ & $11.1$ & $4.4$ & $13.2$ & $6.5$ 
        \\
        LLM-Grounder~\cite{llmgrounder} & ICRA'24 & - & -  & - & - & $17.1$ & $5.3$ 
        \\
        WS-3DVG~\cite{ws3dvg}  & ICCV'23 & - & - & - & - & $27.4$ &  $22.0$ 
        \\
        ZSVG3D~\cite{visualprogramming} & CVPR'24 & $63.8$ & $58.4$ & $27.7$ & $24.6$ & $36.4$ & $32.7$
        \\
        3D-GRAND~\cite{3d-grand} & CVPR'25 &  $54.4$ & $36.4$ & $26.0$ & $20.8$ & $38.0$ & $27.4$\\
        \textbf{Reason3DVG-8B} & Ours & \textbf{76.6} & \textbf{69.5} & \textbf{31.1} & \textbf{27.3} & \textbf{38.7} & \textbf{34.4} \\
        \bottomrule
    \end{tabular}}
    \label{tab:scanrefer}
\end{table*}

\subsection{3D Visual Grounding LLM Fine-tuning}
We employ the open-source LLM Llama-3.1-8B~\cite{llama3} as our base model. The base model Llama-3.1-8B is fine-tuned on our collected 3.2K 3D visual grounding reasoning data. The model is supervised by the step-by-step reasoning responses generated by \texttt{GPT-4o}, using the standard cross-entropy loss with the LLM's Next-Token-Prediction~(NTP) objective, which can be formally defined as:

\begin{equation}
\mathcal{L}_{\text{CE}} = - \sum_{t=1}^{T} \log P_\theta(y_t \mid y_{<t}, x)
\end{equation}
where $x$ denotes the input~(3D object proposals in text format and target query), $y_{1:T}$ is the ground-truth response sequence from \texttt{GPT-4o}, and $P_\theta$ is the model's predicted probability distribution over tokens at each time step $t$. Each training sample includes the 3D scene context, a target object query, and a structured four-stage reasoning response. By learning from these detailed reasoning sequences, the model is encouraged not only to predict the correct answer but also to explicitly generate interpretable intermediate reasoning steps.

\subsection{Inference Pipeline}
During test time, we follow previous non-fine-tuned methods~\cite{visualprogramming,li2024seeground,3d-grand} and use 3D object detector Mask3D~\cite{schult2023mask3d} that takes point clouds as input to generate object proposals and corresponding class labels. These proposals are parsed into a structured text format, with each object assigned a unique object identifier ID. The object proposals are fed into our \model~along with the given language query. \model~then conduct structured reasoning and predict the referring target object's ID. An example of the input format and inference prompt can be found in our appendix.
\section{Experiments}
\label{sec:exp}
\subsection{Implementation Details}
We use Llama-3.1-8B~\cite{llama3} as our base model and conduct fine-tuning using our collected 3D visual grounding reasoning data, which consists of 3.2K 3D scenes, target queries, and structured reasoning responses. More details on training hyperparameters and implementation setup can be found in our appendix.

\subsection{Benchmarks}
We evaluated on two large-scale 3D visual grounding benchmarks, including ScanRefer~\cite{scanrefer} and NR3D~\cite{referit3d} following previous works~\cite{visualprogramming,li2024seeground,3d-grand,llmgrounder}. Both benchmarks leverage 3D scenes collected from ScanNet~\cite{scannet}. We follow previous zero-shot methods~\cite{visualprogramming,li2024seeground,3d-grand} and use Mask3D~\cite{schult2023mask3d} to obtain the same 3D bounding boxes and the object class label for consistent evaluation and fair comparison.

\begin{table*}[!t]
    \centering
    \small
    \caption{Performance on NR3D. Queries are labeled as \textit{``Easy"} (one distractor) or \textit{``Hard"} (multiple distractors), and as 
    \textit{``View-Dependent"} or \textit{``View-Independent"} based on viewpoint requirements for grounding. \textsuperscript{\ding{61}}We list the performance using oracle class label to provide reference for the upper bound performance.}
    \resizebox{\linewidth}{!}{
    \begin{tabular}{l|c|cccc|c}
        \toprule
        \textbf{Method} & \textbf{Source} & 
        \textbf{Easy} & \textbf{Hard} & \textbf{Dep.} & \textbf{Indep.} & \textbf{Overall} 
        \\
        \midrule
        \textit{\textbf{Fine-tuned on Training Set}} \\
        ReferIt3DNet~\cite{referit3d} & ECCV'20 & $43.6$ & $27.9$ & $32.5$ & $37.1$ & $35.6$ \\
        TGNN~\cite{huang2021text} & AAAI'21 & $44.2$ & $30.6$ & $35.8$ & $38.0$ & $37.3$ \\
        3DRefTransformer~\cite{abdelreheem20223dreftransformer} & WACV'22 & $46.4$ & $32.0$ & $34.7$ & $41.2$ & $39.0$ \\
        InstanceRefer~\cite{yuan2021instancerefer} & ICCV'21 & $46.0$ & $31.8$ & $34.5$ & $41.9$ & $38.8$ \\
        FFL-3DOG~\cite{FFL-3DOG} & ICCV'21 & $48.2$ & $35.0$ & $37.1$ & $44.7$ & $41.7$ \\
        LanguageRefer~\cite{roh2022languagerefer} & CoRL'22 & $51.0$ & $36.6$ & $41.7$ & $45.0$ & $43.9$ \\
        3DVG-Transformer~\cite{zhao20213dvg} & ICCV'21 & $48.5$ & $34.8$ & $34.8$ & $43.7$ & $40.8$ \\
        TransRefer3D~\cite{he2021transrefer3d} & MM'21 & $48.5$ & $36.0$ & $36.5$ & $44.9$ & $42.1$ \\
        BUTD-DETR~\cite{butd-detr} & ECCV'22 &  $60.7$ & $48.4$ & $46.0$ & $58.0$ & $54.6$ \\
        \midrule
        \textit{\textbf{Not Fine-tuned on Training Set}} \\
        WS-3DVG~\cite{ws3dvg} & ICCV'23 & $27.3$ & $18.0$ & $21.6$ & $22.9$ & $22.5$ \\
        ZSVG3D~\cite{visualprogramming} & CVPR'24 & $46.5$ & $\textbf{31.7}$ & $36.8$ & $40.0$ & $39.0$ \\
        SeeGround~(InternVL2-8B)~\cite{li2024seeground} & CVPR'25 & $43.6$ & $25.8$ & $32.6$ & $35.4$ & $34.3$ \\
        SeeGround~(InternVL2-26B)~\cite{li2024seeground} & CVPR'25 & $46.8$ & $29.8$ & $34.7$ & $39.8$ & $38.0$ \\
        \textbf{Reason3DVG-8B} & Ours & \textbf{50.4} & 31.0 & \textbf{37.3} & \textbf{42.0} & \textbf{40.4} \\

        \midrule

        \textbf{Reason3DVG-8B}\textsuperscript{\ding{61}} & Ours & \textbf{63.0} & \textbf{36.7} & \textbf{40.1} & \textbf{54.5} & \textbf{49.3} \\
        \bottomrule
    \end{tabular}
    }
    \label{tab:nr3d}
\end{table*}

\paragraph{ScanRefer.} ScanRefer provides 51,500 natural language descriptions from 800 different 3D scenes. We follow previous works~\cite{visualprogramming,llmgrounder,butd-detr} and adopt the validation set of ScanRefer for evaluation, which contains 9,508 language queries. We report ScanRefer's standard evaluation metrics: Accuracy@0.25 and Accuracy@0.5, where 0.25 and 0.5 are different IoU thresholds of 3D bounding boxes.

\paragraph{NR3D.} NR3D contains 41.5K natural language captions collected by humans, featuring multiple spatial relationships. Unlike ScanRefer, NR3D provides the GT 3D bounding boxes in the scene along with their corresponding object IDs. Given a language query, the visual grounding model is required to predict the target object ID that the given query refers to. We adhere to the original NR3D benchmark's setup, categorizing text queries into "Easy" (scenarios with only one same-class distractor) and "Hard" (scenarios with multiple same-class distractors). Additionally, queries are also classified as "View-Dependent" or "View-Independent," depending on whether specific viewpoints are necessary for correctly identifying the target object.

\subsection{Comparison with Non-fine-tuned Methods}
We compare our performance with other SOTA methods on two large-scale 3D visual grounding benchmarks, including ScanRefer~(Tab.\ref{tab:scanrefer}) and NR3D~(Tab.~\ref{tab:nr3d}). On ScanRefer, our model outperforms the recent zero-shot methods ZSVG3D~\cite{visualprogramming} and 3D-GRAND~\cite{3d-grand}. Notably, unlike ZSVG3D, which requires multiple in-context examples and proprietary GPT-4 during inference, our model conducts reasoning and prediction without any in-context examples. Compared with 3D-GRAND, we achieve better performance when training on much less data~(3D-GRAND is trained on \textbf{200K} QA pairs, while our model is only trained on 3.2K QA pairs). On NR3D, our model outperforms recent zero-shot SOTA methods ZSVG3D and SeeGround~\cite{li2024seeground} when using the same bounding box proposals from Mask3D as input. Furthermore, these two methods utilize extra visual input to achieve more fine-grained 3D scene understanding, while our model achieves better results by only using the object bounding box proposals in text format. Incorporating extra object visual features might potentially further improve our method's performance.

\begin{table*}[t]
    \centering
    \small
    \caption{Ablation study on the effectiveness of fine-tuning. We report accuracy (\%) on Easy, Hard, Depth-dependent (Dep.), Depth-independent (Indep.), and Overall splits.}
    \label{tab:training}
    \begin{tabular}{l|c c c c c}
        \toprule
        \textbf{Setup} & \textbf{Easy} & \textbf{Hard} & \textbf{Dep.} & \textbf{Indep.} & \textbf{Overall} \\
        \midrule
        Llama-3.1-8B & $44.6$ & $22.7$ & $31.1$ & $34.5$ & $33.3$ \\
        + Fine-tuning & $\mathbf{63.0}$ $(+18.4)$ & $\mathbf{36.7}$ $(+14.0)$ & $\mathbf{40.1}$ $(+9.0)$ & $\mathbf{54.5}$ $(+20.0)$ & $\mathbf{49.3}$ $(+16.0)$ \\
        \bottomrule
    \end{tabular}
\end{table*}

\begin{table*}[t]
    \centering
    \small
    \caption{Ablation study on the impact of reasoning supervision during fine-tuning on NR3D. We report accuracy (\%) across Easy, Hard, Depth-dependent (Dep.), Depth-independent (Indep.), and Overall splits.}
    \label{tab:reason}
    \begin{tabular}{c | c c c c c}
        \toprule
        \textbf{Reasoning supervision} & \textbf{Easy} & \textbf{Hard} & \textbf{Dep.} & \textbf{Indep.} & \textbf{Overall} \\
        \midrule
         & $42.2$ & $25.4$ & $30.2$ & $35.4$ & $33.5$ \\
        \cmark & $\mathbf{63.0}$ $(+20.8)$ & $\mathbf{36.7}$ $(+11.3)$ & $\mathbf{40.1}$ $(+9.9)$ & $\mathbf{54.5}$ $(+19.1)$ & $\mathbf{49.3}$ $(+15.8)$ \\
        \bottomrule
    \end{tabular}
\end{table*}



\subsection{Comparison with Fine-tuned Methods}
Despite SOTA accuracy among non-fine-tuned methods and outperforming multiple fine-tuned baselines~\cite{referit3d,huang2021text,abdelreheem20223dreftransformer,yuan2021instancerefer}, our results still lag behind SOTA models that are directly fine-tuned on real-world, large-scale data. We investigate the reason for such a gap by using the oracle object class label as input for NR3D, and found that our accuracy can be largely improved and even achieve comparable performance with SOTA fine-tuned models. This suggests that fine-tuned methods can obtain better detection and visual features from the direct supervision of large-scale, real-world data, while our accuracy is limited by the detection quality. Providing the LLM richer semantic clues for reasoning with more sophisticated detectors or captioner~\cite{scan2cap} serves as a promising way to improve the performance.
\subsection{Ablation Studies}

\paragraph{Effectiveness of fine-tuning.}
To show the effectiveness and contribution of our data, we compare the performance on NR3D after fine-tuning. In Tab.~\ref{tab:training}, the model's performance increased by a notable margin across all query categories after fine-tuning, with the overall accuracy increased by 16\%, showing the effectiveness of our data.

\paragraph{Importance of reasoning supervision.}
We trained the base model on our collected data without the reasoning process to verify the importance of reasoning supervision. In this training setting, the model directly generates the final prediction without the reasoning process. This training strategy follows existing 3D LLM methods like 3D-GRAND~\cite{3d-grand} and 3D-LLM~\cite{3dllm}, which directly fine-tune the LLM with short and concise answers without reasoning process. Table.~\ref{tab:reason} showcases that involving the reasoning process during training can largely improve the performance, which demonstrates the importance of reasoning.

\begin{table}[!t]
    \centering
    \small
    \caption{Ablation on domain generalization after fine-tuning. We report accuracy (\%) on In-Domain and Out-of-Domain queries.}
    \label{tab:domain}
    \begin{tabular}{l | c c}
        \toprule
        \textbf{Model} & \textbf{In-Domain} & \textbf{Out-of-Domain} \\
        \midrule
        Llama-3.1-8B & $34.5$ & $32.1$ \\
        + Fine-tuning & $\mathbf{49.6}$ $(+15.1)$ & $\mathbf{49.1}$ $(+17.0)$ \\
        \bottomrule
    \end{tabular}
\end{table}

\paragraph{Generalization to out-of-domain queries.} Our collected data includes only seven common in-domain spatial relationships; the goal is not to mimic the real-world complexity, but some simple data to supervise the LLM to think step-by-step in the 3DVG task. To verify the LLM's generalization ability on unseen and complex real-world queries, we examine the grounding accuracy improvement of our model on in-domain and out-of-domain queries after fine-tuning on our collected data. We found that despite the distribution shift to the complex real-world and unseen, novel spatial relationship, our model still demonstrates strong generalization ability, with more than 15\% performance improvements on both in-domain and out-of-domain unseen queries, as shown in Tab.~\ref{tab:domain}.

\paragraph{Fine-tuning with varied data scale.}
In our experiments, we found that training the LLM on a subset of our data can achieve comparable performance when fine-tuned for more iterations. This result confirms our hypothesis that recent LLMs like Llama-3 already possess basic reasoning abilities. Fine-tuning on our collected small-scale data benefits the LLM’s capability to reason step-by-step following our four-stage format, which improves the LLM's accuracy on 3D visual grounding task and further enables it to generalize to unseen spatial relationships.

\subsection{Qualitative Results}
\begin{figure*}[!t]
    \centering
    \includegraphics[width=\linewidth, height=500pt]{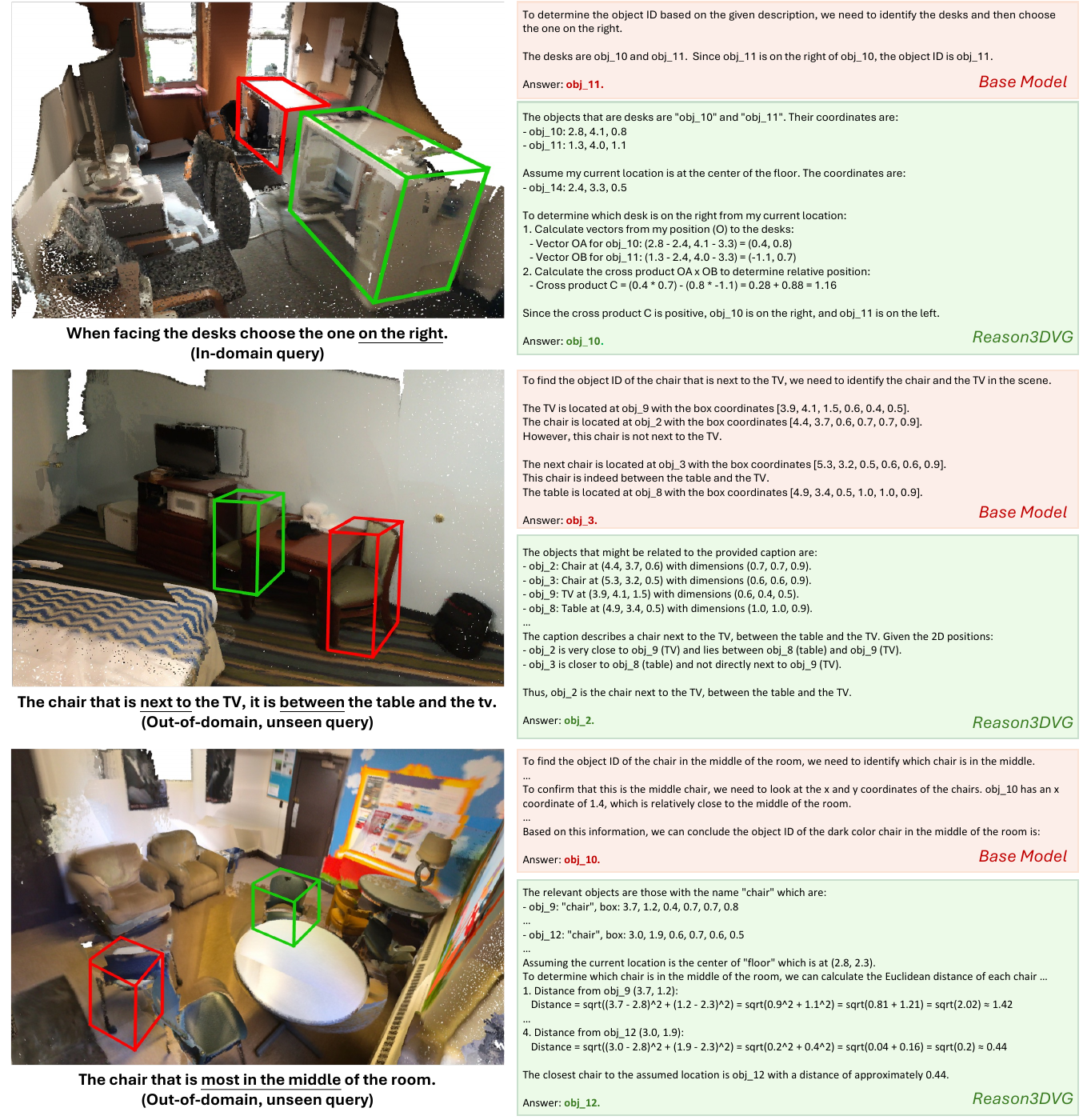}
    \caption{Qualitative results from NR3D. Green boxes and red boxes indicate predictions from our \model~and the base model, respectively. All predicted bounding boxes are re-plotted for better clarity.}
    \label{fig:qual}
\end{figure*}

We provide grounding qualitative results in Fig.~\ref{fig:qual}. We include multiple results from in-domain query, which has similar spatial relationships in our training data, and two out-of-domain queries that have unseen and more complex spatial relationships. The qualitative examples demonstrate the effectiveness of our fine-tuning strategy to enable LLM to conduct 3D visual grounding on the complex language queries and 3D scenes in real-world scenarios.

\section{Limitations and Future Work}
Despite our SOTA results, our accuracy is primarily limited by the quality of object proposals generated by the 3D object detector. Incorporating better detectors and object captioner with richer semantic information can enhance the LLM's reasoning process. 


\section{Conclusion}
\label{sec:conclusion}
In this work, we propose an automatic data pipeline for generating 3D visual grounding data with reasoning supervision, requiring no human annotations. Using our collected data, we perform LLM fine-tuning and introduce \model, a strong 3D visual grounding LLM. \model~outperforms multiple SOTA proprietary LLM-based methods on ScanRefer and NR3D, and surpasses 3D-GRAND using only 1.6\% of its training data. Our results show that detailed reasoning supervision, not just data scale, is key to advancing 3D visual grounding with LLMs.

\clearpage

{
    \small
    \bibliographystyle{ieeenat_fullname}
    \bibliography{main}
}

\clearpage
\setcounter{page}{1}
\maketitlesupplementary

\noindent The supplementary material is structured as follows:

\begin{itemize}[leftmargin=7.5mm]
\setlength{\itemsep}{2pt}
\item LLM training configuration in Section. ~\ref{sec:config}.

\item Scene generation details in Section.~\ref{sec:data_scene}.
\item Training data statistics in Section.~\ref{sec:data_stats}.
\item Prompts for data collection in Section.~\ref{sec:data_prompt}.
\item Spatial relationship template in Section~\ref{sec:template}.

\end{itemize}

\section{Training configurations}
\label{sec:config}

In this part, we provide our training configurations and hyperparameter settings. We conduct post-training on Llama-3.1-8B-Instruct~\cite{llama3}, using the llama-cookbook framework with the configurations listed in Table~\ref{tab:training_parameters}.

\begin{table}[H]
\centering
\caption{Training configurations}
\label{tab:training_parameters}
\begin{tabular}{|l|c|}
\hline
\textbf{Parameter} & \textbf{Value} \\ \hline
    FSDP & enabled \\ \hline
    Learning rate & $1 \times 10^{-5}$ \\ \hline
    Number of epochs & 2 \\ \hline
    Batch size for training & 8 \\ \hline
    Use fast kernels & True \\ \hline
    Run validation & False \\ \hline
    Batching strategy & padding \\ \hline
    Context length & 4096 \\ \hline
    Gradient accumulation steps & 1 \\ \hline
    Gradient clipping & False \\ \hline
    Gradient clipping threshold & 1.0 \\ \hline
    Weight decay & 0.0 \\ \hline
    Gamma & 0.85 \\ \hline
    Seed & 42 \\ \hline
    Use FP16 precision & False \\ \hline
    Mixed precision & True \\ \hline
\end{tabular}
\end{table}

\section{Scene generation details}
\label{sec:data_scene}
We prompt \texttt{GPT-4o} to generate 40 types of common indoor scene objects and their common dimensions in meters to place in the 3D scene during data collection. The full list of objects and their corresponding dimensions can be found in Table.~\ref{tab:objects}. We also apply random variation to the object dimension when placing them in the scene to enrich 3D scene diversity. The 3D scene generation code is attached with the supplementary material for reference.

\begin{table}[t]
\centering
\caption{Scene objects and their corresponding dimensions in (width, length, height) format.}
\label{tab:objects}
\begin{tabular}{|l|c|}
\hline
\textbf{Object} & \textbf{Dimensions (m)} \\ \hline
Window & (0.2, 1.3, 1.2) \\ \hline
Cabinet & (1.0, 0.5, 2.0) \\ \hline
Bed & (2.0, 2.2, 1.0) \\ \hline
Chair & (0.6, 0.6, 1.0) \\ \hline
Sofa & (2.0, 1.0, 1.0) \\ \hline
Table & (1.5, 1.0, 0.75) \\ \hline
Door & (0.9, 0.1, 2.0) \\ \hline
Bookshelf & (1.0, 0.3, 2.0) \\ \hline
Picture & (0.8, 0.05, 0.6) \\ \hline
Counter & (1.5, 0.6, 0.9) \\ \hline
Desk & (1.4, 0.8, 0.75) \\ \hline
Curtain & (2.0, 0.1, 2.0) \\ \hline
Refrigerator & (0.8, 0.8, 1.8) \\ \hline
TV & (1.0, 0.1, 0.6) \\ \hline
Trash Can & (0.4, 0.4, 0.7) \\ \hline
Microwave & (0.6, 0.5, 0.4) \\ \hline
Oven & (0.7, 0.6, 0.9) \\ \hline
Toaster & (0.3, 0.2, 0.3) \\ \hline
Mirror & (1.0, 0.05, 1.5) \\ \hline
Clock & (0.15, 0.1, 0.1) \\ \hline
Mug & (0.08, 0.1, 0.08) \\ \hline
Smartphone & (0.08, 0.15, 0.01) \\ \hline
Wallet & (0.12, 0.08, 0.02) \\ \hline
Remote & (0.05, 0.18, 0.02) \\ \hline
Mouse & (0.07, 0.04, 0.12) \\ \hline
Keyboard & (0.18, 0.05, 0.1) \\ \hline
Book & (0.15, 0.02, 0.22) \\ \hline
Pen & (0.015, 0.015, 0.14) \\ \hline
Light Bulb & (0.06, 0.1, 0.06) \\ \hline
Headphones & (0.18, 0.15, 0.1) \\ \hline
Glasses & (0.14, 0.05, 0.02) \\ \hline
Candle & (0.07, 0.15, 0.07) \\ \hline
Soap Bar & (0.1, 0.02, 0.06) \\ \hline
Spoon & (0.04, 0.02, 0.16) \\ \hline
Fork & (0.03, 0.02, 0.18) \\ \hline
USB & (0.07, 0.02, 0.02) \\ \hline
Dice & (0.02, 0.02, 0.02) \\ \hline
Key & (0.05, 0.02, 0.12) \\ \hline
Coin & (0.03, 0.003, 0.03) \\ \hline
\end{tabular}
\end{table}

\clearpage

\begin{table}[t]
    \centering
    \caption{Training data statistics.}
    \resizebox{0.98\linewidth}{!}{
    \begin{tabular}{c|ccccccc}
    \toprule
    Relationship & Closest & Farthest & Next to & Left & Right & Largest & Smallest\\
    \midrule
    \# of data & 460 & 448 & 458 & 430 & 411 & 479 & 481\\
    \bottomrule
    \end{tabular}
    }
    \label{tab:data}
\end{table}

\section{Training data statistics}
\label{sec:data_stats}
We created a total of 3,500 3D scenes along with corresponding queries, with 700 for each spatial relationship. After filtering out incorrect prediction and wrong format responses from GPT-4o~\cite{4o}, we retained 3,167 data points for training, see Table.~\ref{tab:data}. 

\section{Spatial relationship templates}
\label{sec:template}
\begin{figure*}
    \centering
    \includegraphics[width=0.48\linewidth]{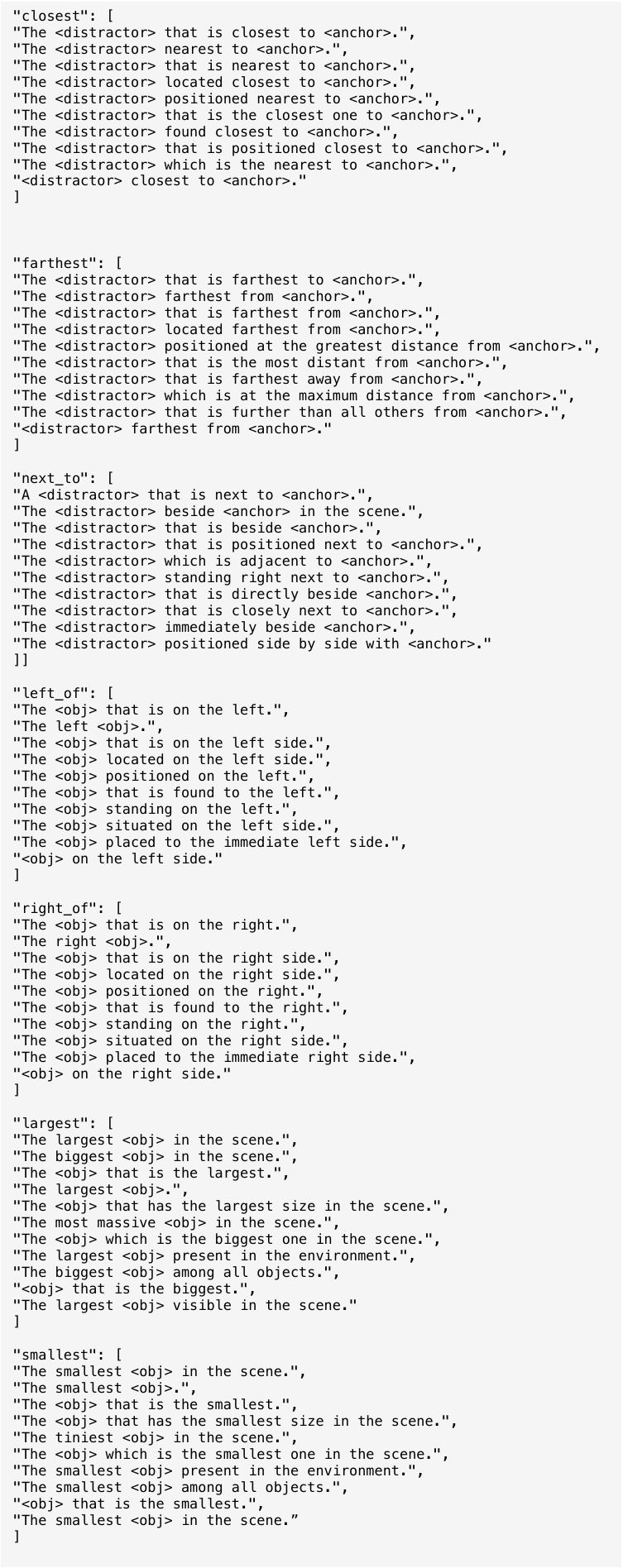}
    \caption{The spatial relationship templates used in our data collection framework.}
    \label{fig:template}
\end{figure*}

We present the spatial relationship templates in Fig.~\ref{fig:template}, which we used to parse the object class name in our data collection framework. We prompt GPT-4o to generate 7 types of template for each spatial relationship.

\section{Data collection and inference prompt}
\begin{figure*}
    \centering
    \includegraphics[width=0.98\linewidth]{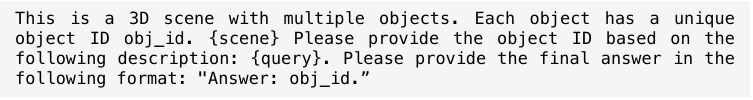}
    \caption{The LLM inference prompt used in our evaluation. {scene} will be filled with the object proposals. An example of object proposals format is in Fig.~\ref{fig:scene}.}
    \label{fig:inference_prompt}
\end{figure*}

\begin{figure*}[t]
    \centering
    \includegraphics[width=0.98\linewidth]{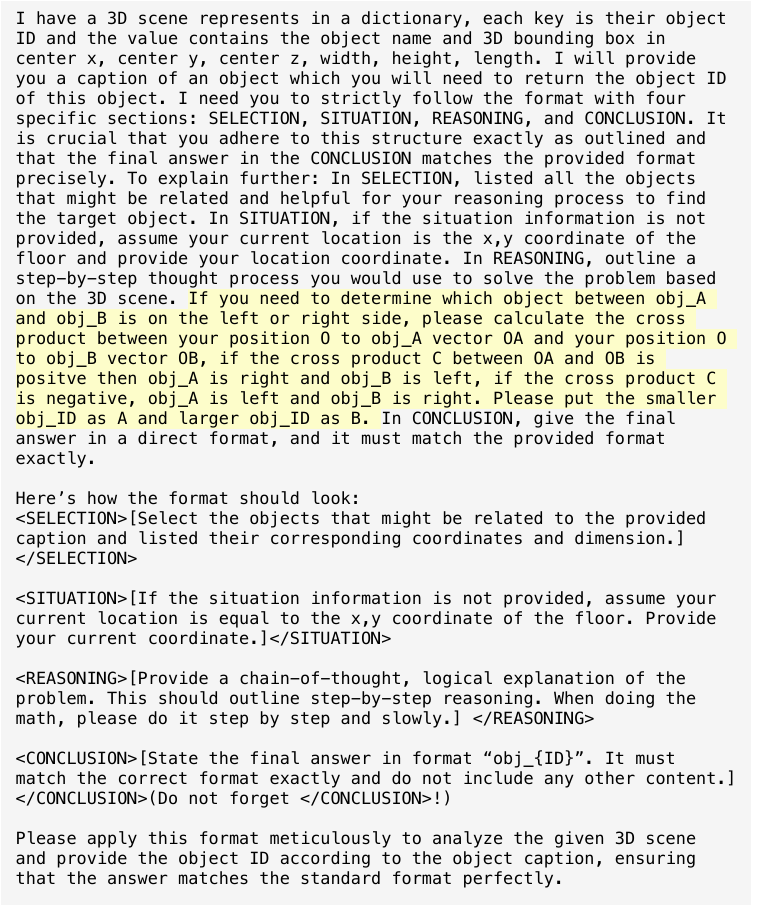}
    \caption{The reasoning data collection prompt that is used to collect reasoning data from GPT-4o. We highlight our designed prompt for reducing spatial relationship hallucination in yellow.}
    \label{fig:4o_prompt}
\end{figure*}

We present our reasoning data collection prompt in Fig.~\ref{fig:4o_prompt}. We highlight our extra prompt in yellow, which can largely reduce the spatial hallucination of LLM. We also present the inference prompt for our \model~in Fig.~\ref{fig:inference_prompt}, which is used for evaluation on NR3D~\cite{referit3d} and ScanRefer~\cite{scanrefer}.
\label{sec:data_prompt}


\clearpage

\begin{figure*}
    \centering
    \includegraphics[width=0.98\linewidth]{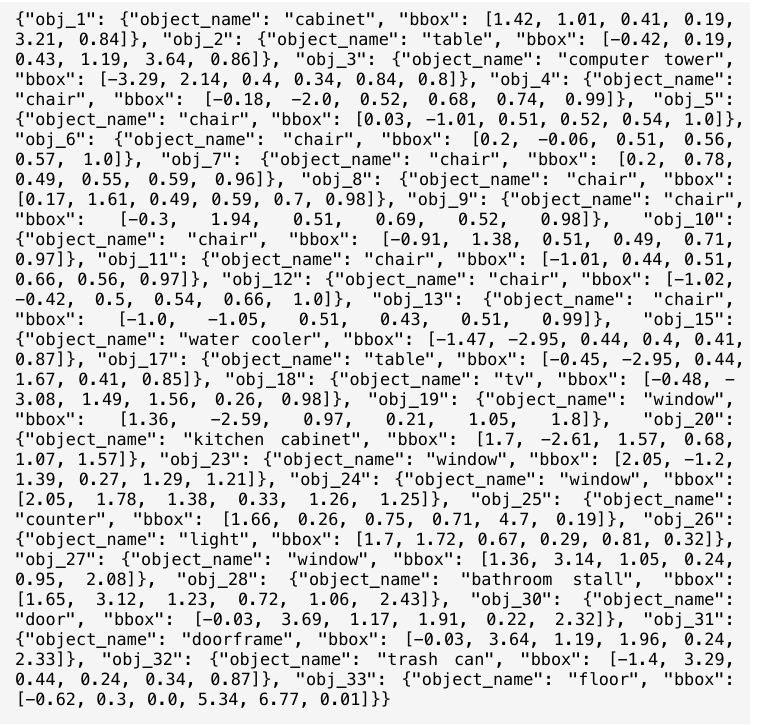}
    \caption{An example of 3D scene format from NR3D, scene0011\_00.}
    \label{fig:scene}
\end{figure*}
\end{document}